\documentclass{article}

\usepackage{graphicx}
\usepackage{subcaption}
\usepackage{amsmath}
\usepackage{amssymb}
\usepackage{multirow}
\usepackage{hyperref}

\usepackage[preprint]{neurips_2023}

\usepackage{natbib}

\usepackage[utf8]{inputenc} 
\usepackage[T1]{fontenc}    
\usepackage{hyperref}       
\usepackage{url}            
\usepackage{booktabs}       
\usepackage{amsfonts}       
\usepackage{nicefrac}       
\usepackage{microtype}      
\usepackage{xcolor}         

\title{Language Model Training Paradigms for Clinical Feature Embeddings}

\author{%
  Yurong Hu\textsuperscript{1}, Manuel Burger\textsuperscript{2}, Gunnar Rätsch\textsuperscript{2}, Rita Kuznetsova\textsuperscript{2} \\
  \texttt{\{yurohu, burgerm, raetsch, mkuznetsova\}@ethz.ch} \\
  \textsuperscript{1} Department of Information Technology and Electrical Engineering, ETH Z\"{u}rich \\
  \textsuperscript{2}Department of Computer Science, ETH Z\"{u}rich
 }

\begin{document}
\maketitle
\begin{abstract}
    In research areas with scarce data, representation learning plays a significant role. This work aims to enhance representation learning for clinical time series by deriving universal embeddings for clinical features, such as heart rate and blood pressure. We use self-supervised training paradigms for language models to learn high-quality clinical feature embeddings, achieving a finer granularity than existing time-step and patient-level representation learning. We visualize the learnt embeddings via unsupervised dimension reduction techniques and observe a high degree of consistency with prior clinical knowledge. We also evaluate the model performance on the MIMIC-III benchmark and demonstrate the effectiveness of using clinical feature embeddings. We publish our code online for replication\footnote{\href{https://github.com/yuroeth/icu_benchmarks}{https://github.com/yuroeth/icu\_benchmarks}}.
\end{abstract}

\section{Introduction}
\label{sec:intro}
\vspace{-0.3cm}
The wide adoption of the EHR system has engendered an unprecedented availability of patient data, which serves as a treasure trove for ML researchers. Such data encapsulates a patient's medical trajectory, inclusive of their medical history, diagnoses, laboratory tests, and treatment interventions. Prior ML research \citep{horn2020set,xu2018raim} primarily aimed at the modification of the backbone sequence model, mostly employing supervised training approaches for the prediction of patient-centric problems like in-hospital length-of-stay and mortality rates. Simultaneously, several studies \citep{yue2022ts2vec,tonekaboni2021unsupervised,yeche2021neighborhood} have successfully applied self-supervised learning methodologies for the extraction of time-step level or patient-level embeddings in time series. However, these high-level embeddings are largely confined to the specific datasets upon which they were trained. Furthermore, the pre-training objectives are mostly focused on contrastive loss, resulting in a deficit of exploration concerning other predictive objectives. \citet{horn2020set} and \citet{tipirneni2022self} considered time series as a set of observation triplets and got the feature level embeddings through the aggregation of three embeddings (i.e., time, feature and value). Both works are confined to the regime of set function learning. \citet{tipirneni2022self} also used an auxiliary self-supervision task for training, but we separate the pre-training and fine-tuning stages in our work, which makes it convenient for the unsupervised feature analysis. Other related works are provided in Appendix \ref{apx1}.
\vspace{-0.4cm}
\paragraph{Our Contribution} In this study, we conduct a granular analysis of representation learning for clinical features such as heart rate and blood pressure. We employ self-supervised training paradigms for language models to obtain more universally applicable clinical embeddings. Specifically, we adopt the Continuous Bag of Words (CBOW) model from Word2Vec~\citep{mikolov2013efficient} and the Masked Language Model (MLM) from BERT~\citep{devlin2018bert}. Experimental analysis demonstrates that leveraging clinical feature embeddings can improve the performance on downstream tasks. Additionally, the clinical feature embeddings obtained from the language model pre-training paradigms show a well-structured latent space, from which we can infer established clinical knowledge.

\section{Methods}
\label{methods}
\begin{figure}[htbp]
    \centering
    \begin{subfigure}[b]{\linewidth}
    \includegraphics[width=\linewidth]      {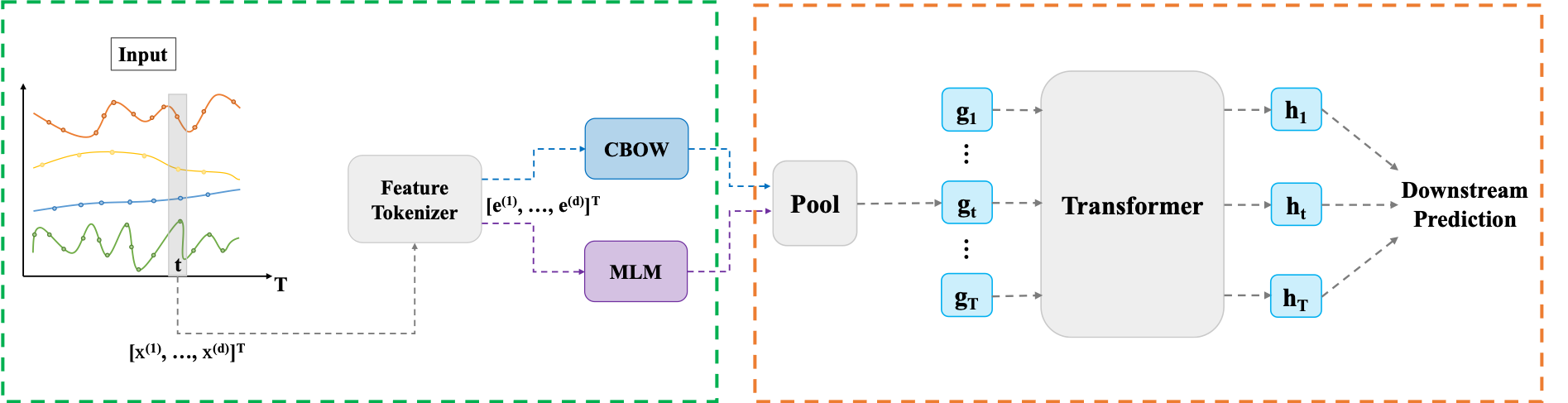}
    \caption{\textbf{Pipeline of our method.} The green box is the pre-training stage, and the orange box represents the fine-tuning stage.}
    \label{fig:pipeline}
    \end{subfigure}
    \hfill
    \begin{subfigure}[b]{0.45\linewidth}
    \includegraphics[width=\linewidth]      {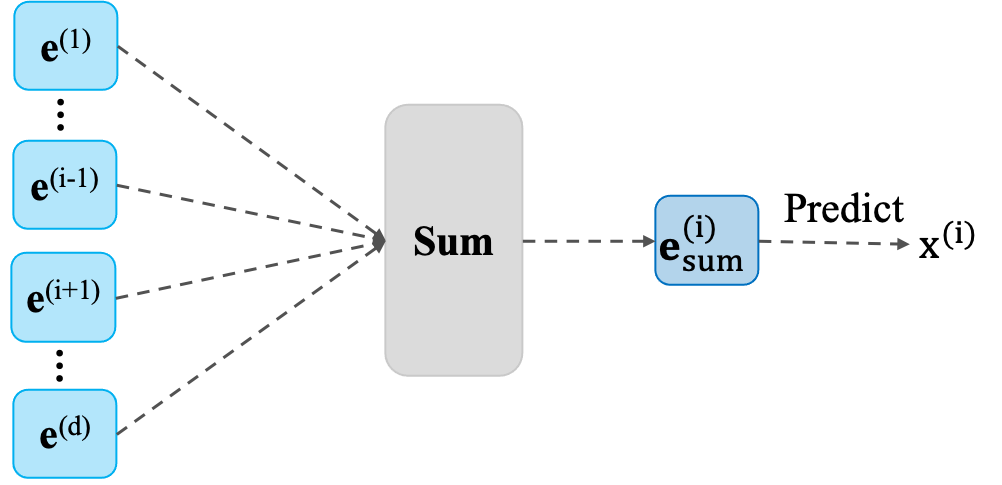}
    \caption{\textbf{CBOW}}
    \label{fig:cbow}
    \end{subfigure}
    \hspace{0.5cm}
    \begin{subfigure}[b]{0.45\linewidth}
    \includegraphics[width=\linewidth]      {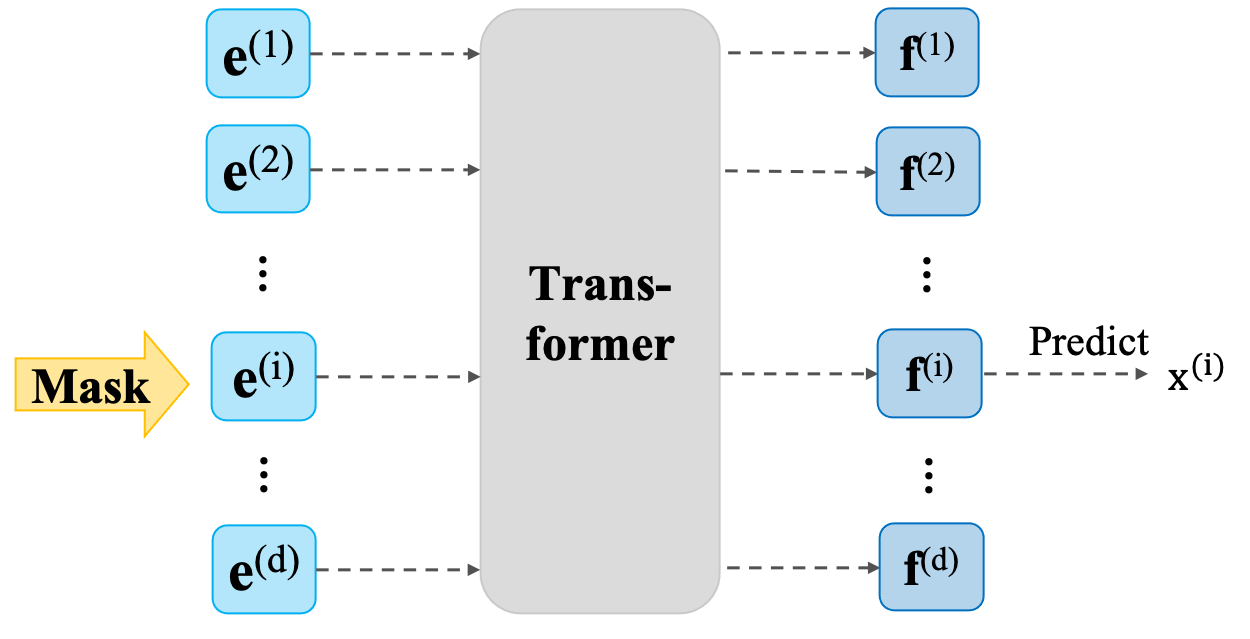}
    \caption{\textbf{MLM}}
    \label{fig:mlm}
    \end{subfigure}
    \caption{Self-supervised learning framework for clinical time series.}
    \label{fig:framework}
\end{figure}

\paragraph{Notations}
We define the whole dataset from ICU patient stay as $\{(\mathbf{X}_i,\mathbf{y}_i) | i=1,2,...,N\}$. Each $\mathbf{X}_i$ is a multivariate time series $\mathbf{X}_i=[\mathbf{x}_{i,1},...,\mathbf{x}_{i,T}]$, where $T$ is the length of the stay $i$. Each time step is $\mathbf{x}_{i,t}=[x_{i,t}^{(1)},...,x_{i,t}^{(d)}]\in \mathbb{R}^d$, where $d$ is the number of clinical features. Depending on the specific task, the label $\mathbf{y}_i$ for patient stay $\mathbf{X}_i$ can be a single value $\mathbf{y}_i \in \mathbb{R}$ that indicates the state of the whole patient stay or a vector $\mathbf{y}_i \in \mathbb{R}^T$ that corresponds to the state of each time step. In the self-supervised learning stage, we consider each time step $\mathbf{x}_{i,t}$ as one sample for the model. For the ease of expression, we omit the subscript and use $\mathbf{x}=[x^{(1)},...,x^{(d)}]$ instead in the following explanation of CBOW and MLM models.
\vspace{-0.4cm}
\paragraph{CBOW}
Given a set of clinical features in a certain time step of patient stay $\mathbf{x}=[x^{(1)},...,x^{(d)}]$, we randomly select one numerical variable $x^{(j)}$ and one categorical variable $x^{(k)}$ to predict. The variables are first fed into the feature tokenizer \citep{gorishniy2021revisiting}, see Appendix \ref{apx1}, which maps discrete feature values to embedding vectors $\mathbf{e}=[\mathbf{e}^{(1)},...,\mathbf{e}^{(d)}]$, where $\mathbf{e}^{(i)}\in \mathbb{R}^{m}$, and $m$ is the embedding dimension. Then we sum the embeddings of all variables except the predicted one, see Figure \ref{fig:cbow}: $\mathbf{e}_{sum}^{(j)}=\sum_{i\neq j}\mathbf{e}^{(i)}$ for predicting value $x^{(j)}$ with the mean squared error $L_{num}=\text{MSE}(\text{Linear}(\mathbf{e}_{sum}^{j}), x^{(j)})$ and $\mathbf{e}_{sum}^{(k)}=\sum_{i\neq k}\mathbf{e}^{(i)}$ for predicting value $x^{(k)}$ with the cross-entropy loss $L_{cat}=\text{CE}(\text{Linear}(\mathbf{e}_{sum}^{k}), x^{(k)})$. Finally we add the two losses together $L=L_{num}+L_{cat}$ for the model update.
\vspace{-0.4cm}
\paragraph{MLM}
 Based on the initial embeddings $\mathbf{e}=[\mathbf{e}^{(1)},...,\mathbf{e}^{(d)}]$ from the Feature Tokenizer \citep{gorishniy2021revisiting}, we randomly mask the embeddings of one numerical variable $\mathbf{e}^{(j)}$ and one categorical variable $\mathbf{e}^{(k)}$, see Figure \ref{fig:mlm}. Similar to MLM in BERT pre-training \citep{devlin2018bert}, we replace the masked positions with (1) the [MASK] embedding $80\%$ of the time (2) a random vector $10\%$ of the time (3) the original feature embedding $10\%$ of the time. The processed embeddings are then passed into the Transformer encoder to get contextual embeddings for each variable $\mathbf{f}=[\mathbf{f}^{(1)},...,\mathbf{f}^{(d)}]$. $\mathbf{f}^{(j)}$ and $\mathbf{f}^{(k)}$ are responsible for predicting the corresponding masked variable value $x^{(j)}$ and $x^{(k)}$ respectively. Similarly, we use the total loss $L=L_{num}+L_{cat}$ for model update, where $L_{num}=\text{MSE}(\text{Linear}(\mathbf{f}^{(j)}), x^{(j)})$, $L_{cat}=\text{CE}(\text{Linear}(\mathbf{f}^{(k)}), x^{(k)})$.
\vspace{-0.4cm}
\paragraph{Downstream Fine-Tuning}
For the downstream task, our input is a patient stay comprising several time steps. The learnt embeddings for clinical features at each time step $t$ are pooled to get the time-step level embeddings: $\mathbf{g}_t=\text{Pool}(\mathbf{e}_{t}^{(1)},...,\mathbf{e}_{t}^{(d)})$ for CBOW and $\mathbf{g}_t=\text{Pool}(\mathbf{f}_{t}^{(1)},...,\mathbf{f}_{t}^{(d)})$ for MLM. Then we feed $\mathbf{g}_t$ into the Transformer encoder to get the contextual embeddings for each time step $[\mathbf{h}_{1},...,\mathbf{h}_{T}]=\text{Transformer}([\mathbf{g}_{1},...,\mathbf{g}_{T}])$. For time-step level predictions, we apply a linear layer for each time step $\hat{y}_{t}=\text{Linear}(\mathbf{h}_{t})$. For stay level predictions, we apply a linear layer to the last time step counted. Subsequently, we compute the task-specific loss function. In our experiments, we use max pooling to get time-step level embeddings and the cross-entropy loss is adopted for both tasks. 
\vspace{-0.3cm}
\section{Experiment Setup}
\vspace{-0.3cm}
\paragraph{Dataset} We use MIMIC-III dataset \citep{johnson2016data} for pre-training and fine-tuning. In self-supervised pre-training, we discard time steps with missing value rate larger than $80\%$ (i.e. more than $15$ missing values out of $18$ features in total). We impute the missing numerical features with the mean and missing categorical features with the mode from the whole dataset. We evaluate the quality of our pre-trained clinical embeddings on two tasks: (1) decompensation and (2) patient mortality at 48 hours after admission from MIMIC-III benchmark \citep{harutyunyan2019multitask}.
\vspace{-0.4cm}
\paragraph{Models} We consider two baseline models. The first model is the Transformer \citep{vaswani2017attention} that takes the raw clinical features as input. The second one is the Feature Tokenizer Transformer (FTT) \citep{gorishniy2021revisiting} which maps the input clinical variables to the embedding vectors before passed to the Transformer encoder. For CBOW and MLM, the feature tokenizer is pre-trained with corresponding self-supervision tasks, as described in Sections \ref{methods}. The detailed training setup and choice of hyper-parameters are shown in Appendix \ref{apx2}.
\vspace{-0.4cm}
\paragraph{Metrics} Given that the downstream tasks are significantly unbalanced classification problems, we use the area under the precision-recall curve (AUPRC) and the area under the receiver operating characteristics curve (AUROC) as the measurement. 
\vspace{-0.3cm}
\section{Results}
\label{sec:results}
\vspace{-0.3cm}
\paragraph{Performance on Downstream Task}
The pre-trained clinical embeddings are evaluated on the decompensation and mortality prediction tasks from the MIMIC-III benchmark \citep{harutyunyan2019multitask}. From Table \ref{tab:mimic_results}, we see that FTT, CBOW and MLM models outperform the Transformer model, demonstrating that feature embeddings are beneficial to clinical predictions. However, CBOW and MLM can not further improve the performance from the FTT model, which suggests that the pre-training of clinical embeddings does not necessarily help the downstream task. However, from the unsupervised feature analysis below, we will see that the pre-trained embeddings have a better connection with prior clinical knowledge than FTT embeddings.
\begin{table}[hbtp]
    \centering
    \begin{tabular}{ccccc}
        \toprule[2pt]
        Task & \multicolumn{2}{c}{Decompensation} &  \multicolumn{2}{c}{Mortality} \\
        \cmidrule(r){2-3} \cmidrule(r){4-5} 
        Metric & AUPRC & AUROC & AUPRC & AUROC \\ \hline
        Transformer & $34.4\pm 0.4$& $91.2\pm 0.1$& $51.5\pm 0.6$& $\mathbf{86.5}\pm 0.3$ \\
        FTT & $\mathbf{36.4}\pm 0.2$ & $\mathbf{91.6}\pm 0.1$ & $\mathbf{53.4}\pm 0.4$ & $85.8\pm 0.1$\\
        CBOW & $36.3\pm 0.4$& $91.4\pm 0.1$& $53.0\pm 0.5$& $85.8\pm 0.3$\\
        MLM & $36.2\pm 0.1$& $\mathbf{91.6}\pm 0.1$& $53.1\pm 0.2$& $86.0\pm 0.2$\\
        \bottomrule[2pt]
    \end{tabular}
    \caption{Performance on two tasks from the MIMIC-III benchmark for different models measured with AUPRC and AUROC. Mean and standard deviation are reported over three runs.}
    \label{tab:mimic_results}
    \vspace{-0.5cm}
\end{table}
\vspace{-0.4cm}
\paragraph{Unsupervised Feature Analysis}
For numerical features, artificial feature values are introduced to the pre-trained feature tokenizer, following which the dimension reduction technique T-SNE \citep{van2008visualizing} is employed on the resultant artificial output to enable visualization.
Given that the authentic input data is standard normalized ($\text{mean}=0,\  \text{std}=1$), we choose $(-3\times \text{std})$, $0$, and $(3\times \text{std})$ as the artificial inputs. Accordingly, low-level feature embeddings are represented as $-3\times W^{num}+b^{num}$ ($\blacktriangledown$), middle-level feature embeddings are $b^{num}$ ($\bullet$), and high-level feature embeddings are $3\times W^{num}+b^{num}$ ($\blacktriangle$). The T-SNE visualizations are depicted in Figure \ref{fig:tsne_num}. The mapping of feature names to their abbreviations is in Appendix \ref{apx2}. On scrutinizing Figure \ref{fig:tsne_num}, we find that several relationships among pre-trained embeddings align with established clinical knowledge. Both in Figure \ref{fig:tsne_num_cbow} and Figure \ref{fig:tsne_num_mlm}, the middle-value feature embeddings ('$\bullet$') tend to cluster (see gray circles), denoting a normal patient state. Further, we observe a proportional relationship among body temperature (Temp), respiratory rate (RR) and heart rate (HR) (see purple rectangles), which agrees with the fact that an increase in body temperature would correspondingly elevate the respiratory rate (RR) and heart rate (HR). Additionally, the MLM embeddings reveal a proportionality between diastolic and systolic blood pressure (DBP \& SBP) (see blue boxes). 
The orange box with low oxygen saturation (OS) and high fractional inspired oxygen ($\text{FiO}_\text{2}$) is a sign of respiratory failure, because OS has positive correlation with partial pressure of oxygen in arterial blood ($\text{PaO}_\text{2}$) and a low P/F ($\text{PaO}_\text{2} / \text{FiO}_\text{2}$) ratio indicates high risk of respiratory failure \citep{huser2024comprehensive}.
These correlations are not observed in the embeddings from FTT, see Appendix \ref{apx3}.
For categorical features, we directly use $W^{cat}+b^{cat}$ as features of varying levels and the results are in Appendix \ref{apx3}, where we also present the results for other ablation studies.
\begin{figure}[htbp]
    \centering
    \begin{subfigure}[b]{.65\linewidth}
    \includegraphics[width=\linewidth]      {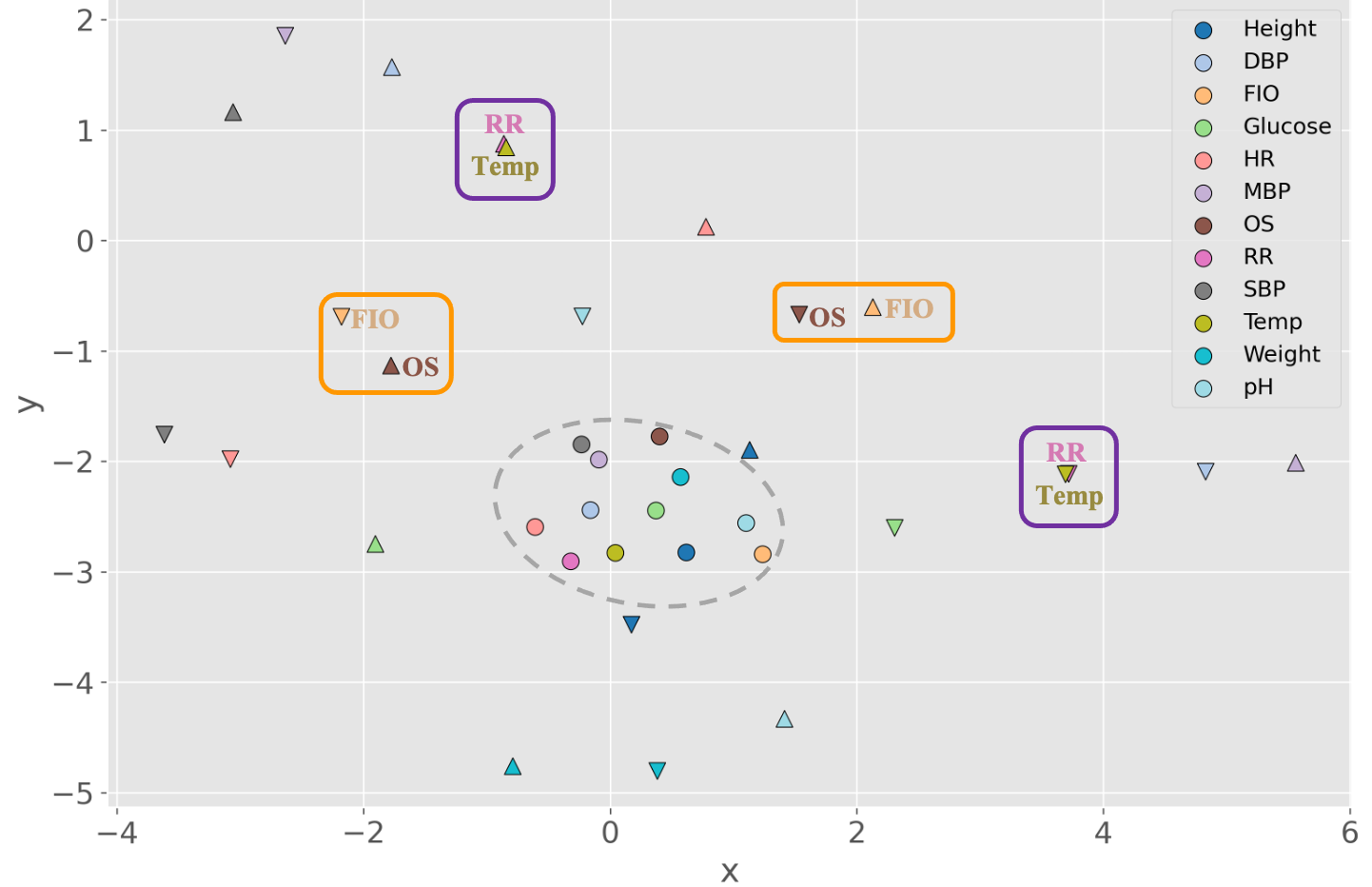}
    \caption{\textbf{CBOW}}
    \label{fig:tsne_num_cbow}
    \end{subfigure}
    \begin{subfigure}[b]{.65\linewidth}
    \includegraphics[width=\linewidth]      {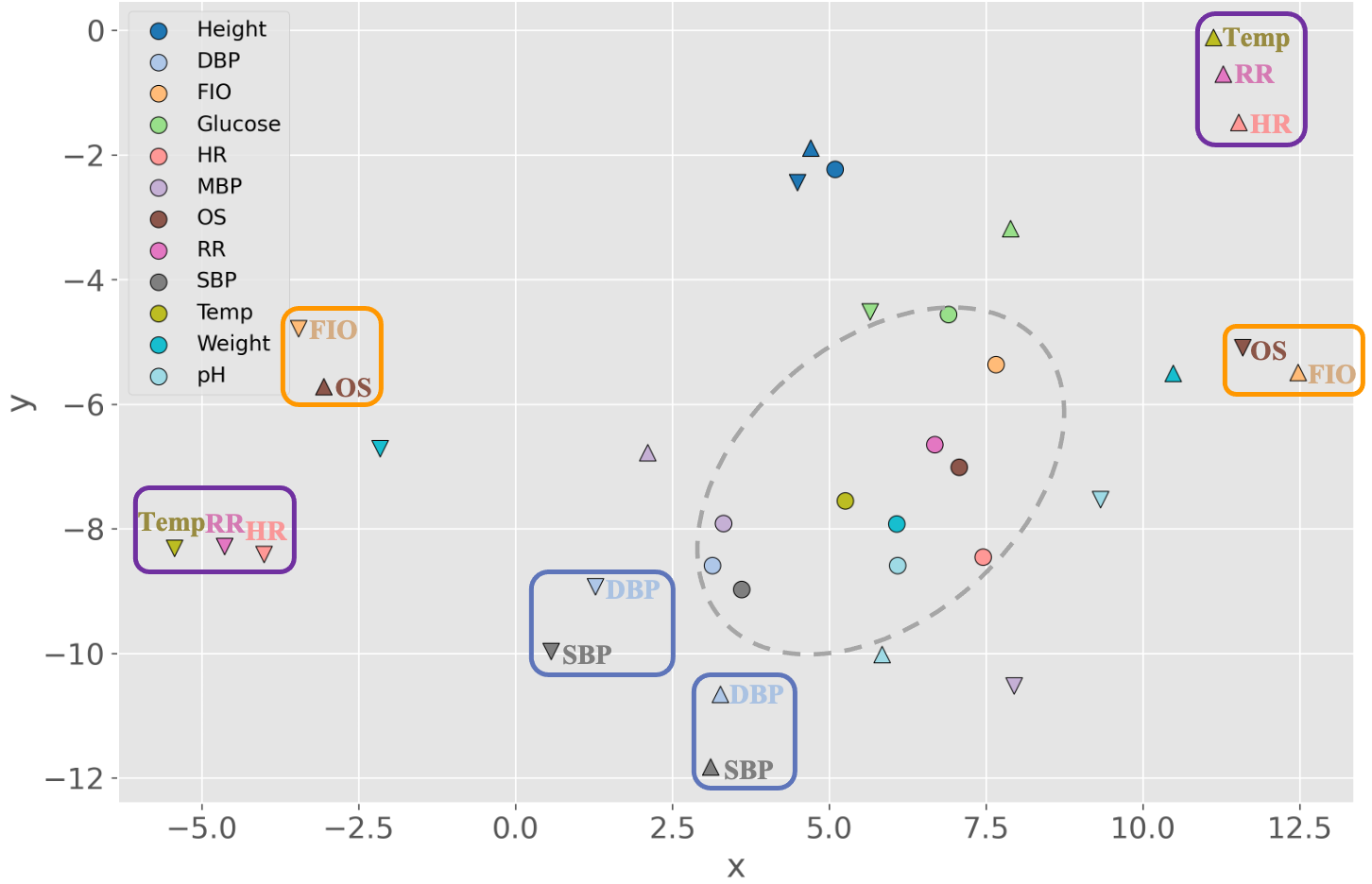}
    \caption{\textbf{MLM}}
    \label{fig:tsne_num_mlm}
    \end{subfigure}
    \caption{T-SNE visualization, with the perplexity value set to $15$,  of numerical feature embeddings from CBOW and MLM (FTT can be found in Appendix \ref{apx3}). Different colors designate the individual features and shapes their magnitude as explained in Section \ref{sec:results}.}
    \label{fig:tsne_num}
\end{figure}
\vspace{-0.8cm}
\section{Conclusion}
\vspace{-0.3cm}
This work seeks to address the challenges faced in representation learning for clinical time series. Existing works are mainly targeted at learning time-step level or patient level feature representations, with a predominant focus on contrastive losses. While improving predictive performance on downstream tasks, these high-level representations suffer from the confinement to specific datasets they were trained on. In an attempt to improve the universality and applicability of clinical feature representations, our study embarks on a granular analysis of representation learning for clinical features. The primary contributions of our work thus include the derivation of universal clinical feature embeddings via CBOW and MLM models, evaluation and analysis of the pre-trained embeddings, and verification of their effectiveness in performance improvement on downstream tasks and interpetability. We believe our findings will encourage further works in exploring the design and application of clinical embeddings.

\newpage
\bibliographystyle{plainnat}
\bibliography{references}

\newpage
\appendix

\section{Related Work}
\label{apx1}
\citet{mikolov2013efficient} devised two distinctive models that enable the projection of words into a continuous vector space. The Continuous Bag of Words (CBOW) model infers the central word using its surrounding context, whereas the skip-gram model forecasts the neighboring words given the central word. Both models proficiently yield superior quality word representations, effectively capturing both syntactic and semantic word similarities. 
Subsequent to the advent of the Transformer model \citep{vaswani2017attention}, \citet{devlin2018bert} introduced BERT, a language representation model developed on the core framework of the Transformer encoder. BERT is engineered to generate deep bidirectional embeddings from a substantial volume of unlabeled text data. Its pre-training tasks include Masked Language Modeling (MLM) and Next Sentence Prediction (NSP). The MLM task prompts the model to predict the masked token within its left and right context, while NSP instructs the model to determine the adjacency of two sentences.

Many studies on time series representation learning focus on obtaining time-step level, stay level or patient level embeddings. 
\citet{tonekaboni2021unsupervised} proposed Temporal Neighborhood Coding (TNC) which leverages the local smoothness inherent to the generative process of time series. They devised a contrastive objective aimed at distinguishing between neighborhood and distant signals.
In later work, \citet{yue2022ts2vec} proposed TS2Vec, a contrastive learning framework for learning representations of time series data in a hierarchical manner. TS2Vec can apply temporal contrast and instance contrast arbitrarily at each layer of the dilated CNN model. 
Besides, \citet{yeche2021neighborhood} designed a Neighborhood Contrastive Learning (NCL) framework for online patient monitoring. NCL incorporates data augmentation techniques for time series data and a novel contrastive objective, which consists of a Neighbor Alignment objective and a Neighbor Discriminative objective. The patient stay level embeddings learnt in this manner proved effective on the MIMIC benchmark and the Physionet 2019 dataset. 

For the acquisition of meaningful representations for clinical features, the initial step is to transform each individual variable into embedding vectors. Gorishniy et al. \citet{gorishniy2021revisiting} summarized the commonly used model architectures for such vectorization, encompassing MLP, ResNet, and Feature Tokenizer + Transformer (FT-Transformer). FT-Transformer first projects numerical and categorical features onto the embedding space respectively, followed by applying a stack of Transformer layers to the embeddings.
Another line of work leverages set function learning for time series \citep{horn2020set, tipirneni2022self}. The set function representation addresses the prevalent issues in time series data such as missing information and irregular time intervals. \citet{tipirneni2022self} treat time series as a set of observation triplets, defined as (time, feature, value). They developed a novel Continuous Value Embedding (CVE) mechanism to embed time and value in the triplet. They also applied self-supervised learning with forecasting as the predictive objective to learn robust feature-level representations. 

\section{Model Parameters and Experiment Setup}
\paragraph{Training setup}
The pipeline of our method is shown in Figure \ref{fig:pipeline}. In the pre-training stage, we use the self-supervised objectives CBOW or MLM to learn feature level embeddings. In the fine-tuning stage, we pool the pre-trained feature level embeddings to get the time-step level embeddings, which are then encoded by the Transformer model for clinical prediction.
\label{apx2}
\begin{figure}[htbp]
    \centering
     \includegraphics[width=0.8\linewidth]{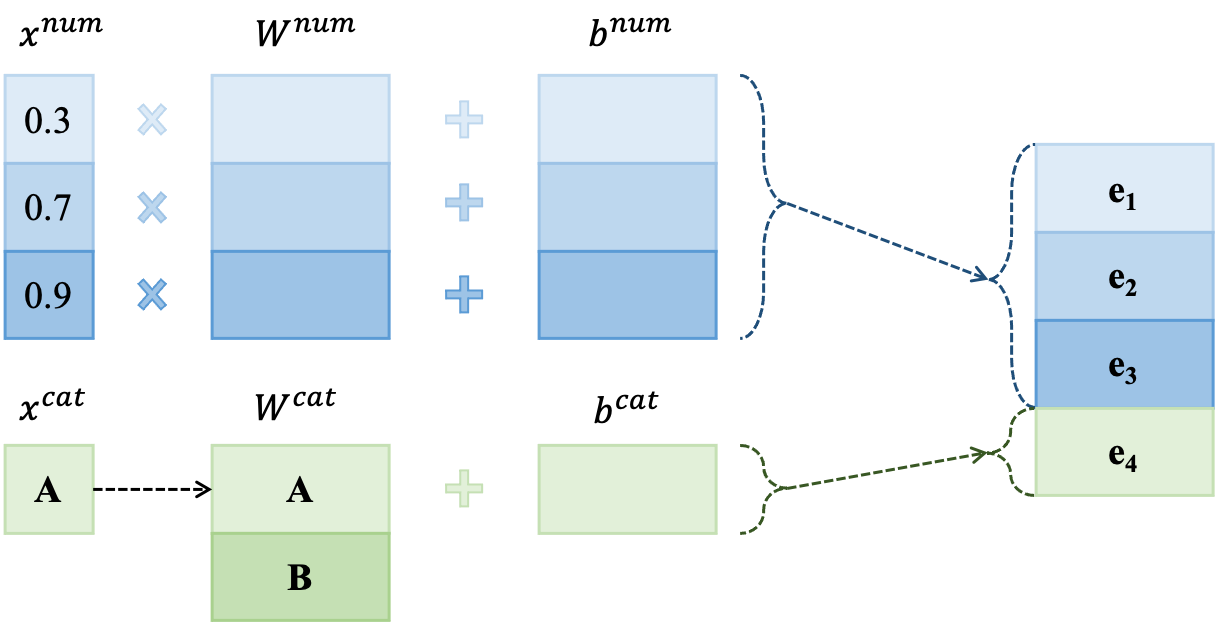}
     \caption{Feature Tokenizer model from \citet{gorishniy2021revisiting}}
     \label{fig:ft}
 \end{figure}

 \begin{table}[htbp]
    \centering
    \begin{tabular}{ccccccc}
         \toprule[2pt]
         batch size & LR & feature\_dim & depth & heads & AUPRC & AUROC \\
         \midrule[1pt]
         8 & 0.001 & 64 & 1 & 1 & $36.6\pm 0.5$ & $89.7\pm 0.1$\\
         8 & 0.001 & 64 & 2 & 2 & $35.8\pm 0.1$ & $89.5\pm 0.1$\\
         8 & 0.001 & 128 & 1 & 1 & $36.1\pm 0.2$ & $89.5\pm 0.1$\\
         8 & 0.001 & 128 & 2 & 2 & $34.7\pm 0.2$ & $89.3\pm 0.1$\\
         8 & 0.0001 & 64 & 1 & 1 & $37.9\pm 0.2$ & $90.1\pm 0.1$\\
         8 & 0.0001 & 64 & 2 & 2 & $37.9\pm 0.2$ & $90.1\pm 0.1$\\
         8 & 0.0001 & 128 & 1 & 1 & $37.8\pm 0.4$ & $90.1\pm 0.1$\\
         8 & 0.0001 & 128 & 2 & 2 & $37.7\pm 0.6$ & $90.0\pm 0.3$\\
         16 & 0.001 & 64 & 1 & 1 & $37.0\pm 0.6$ & $89.9\pm 0.1$\\
         16 & 0.001 & 64 & 2 & 2 & $36.8\pm 0.3$ & $89.7\pm 0.1$\\
         16 & 0.001 & 128 & 1 & 1 & $36.5\pm 0.1$ & $89.8\pm 0.1$\\
         16 & 0.001 & 128 & 2 & 2 & $35.0\pm 1.0$ & $89.5\pm 0.3$\\
         16 & 0.0001 & 64 & 1 & 1 & $37.8\pm 0.1$ & $90.1\pm 0.1$ \\
         16 & 0.0001 & 64 & 2 & 2 & $37.9\pm 0.2$ & $90.2\pm 0.1$ \\
         16 & 0.0001 & 128 & 1 & 1 & $\textbf{38.4}\pm 0.1$ & $\textbf{90.3}\pm 0.0$\\
         16 & 0.0001 & 128 & 2 & 2 & $38.0\pm 0.4$ & $90.1\pm 0.1$\\
         \bottomrule[2pt]
    \end{tabular}
    \caption{Random search results for fine-tuning parameters. AUPRC and AUROC is on the validation set of MIMIC-III decompensation prediction task. We report mean and standard deviation from three runs.}
    \label{tab:rs_ft}
\end{table}

\begin{table}[htbp]
    \centering
    \begin{tabular}{ccc}
        \toprule[2pt]
        Module & Parameter & Value \\
        \hline
        \multirow{6}{*}{Fine-Tune} & batch size & $16$\\
                                   & learning rate & $0.0001$\\
                                   & feature dimension & $128$\\
                                   & depth & $1$\\
                                   & num\_heads & $1$\\
                                   & pooling & max\\
        \hline
        \multirow{3}{*}{CBOW} & batch size & $256$\\
                              & learning rate & $0.01$\\
                              & feature dimension & $256$\\
        \hline
        \multirow{5}{*}{MLM} & batch size & $512$\\
                             & learning rate & $0.0001$\\
                             & feature dimension & $128$\\      
                             & depth & $2$\\
                             & num\_heads & $1$\\
        \bottomrule[2pt]
    \end{tabular}
    \caption{Training Hyper-parameters.}
    \label{tab:parameters}
\end{table}

\begin{table}[htbp]
    \centering
    \begin{tabular}{cc}
        \toprule[2pt]
         Abbr. & Name \\
         \midrule[1pt]
         DBP & diastolic blood pressure \\
         FIO & fraction of inspired oxygen \\
         HR & heart rate \\
         MBP & mean blood pressure \\
         OS & oxygen saturation \\
         RR & respiratory rate \\
         SBP & systolic blood pressure \\
         Temp & temperature \\
         CRR & capillary refill rate \\
         GCST & Glascow coma scale total \\
         GCSEO & Glascow coma scale eye opening \\
         GCSMR & Glascow coma scale motor response \\
         GCSVR & Glascow coma scale verbal response \\ 
         \bottomrule[2pt]
    \end{tabular}
    \caption{Clinical feature name abbreviations.}
    \label{tab:correspondence}
\end{table}

\newpage
\section{Additional Results}
\label{apx3}

\begin{figure}
    \centering
    \includegraphics[width=0.5\linewidth]{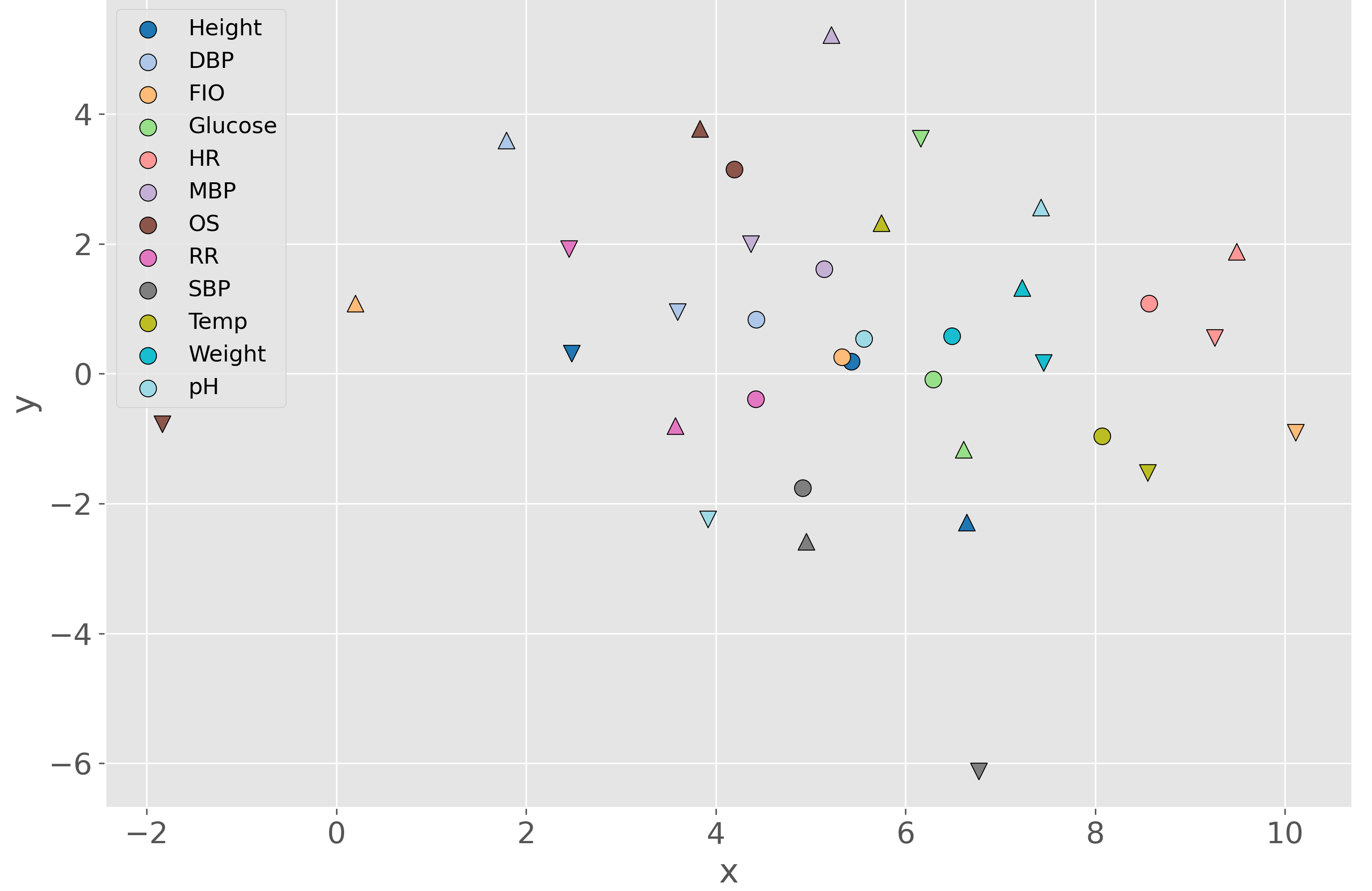}
    \caption{T-SNE visualization, with the perplexity value set to $15$, of numerical feature embeddings from FTT.}
    \label{fig:tsne_num_ftt}
\end{figure}

\begin{figure}[htbp]
    \centering
    \begin{subfigure}[b]{0.45\linewidth}
    \includegraphics[width=\linewidth]      {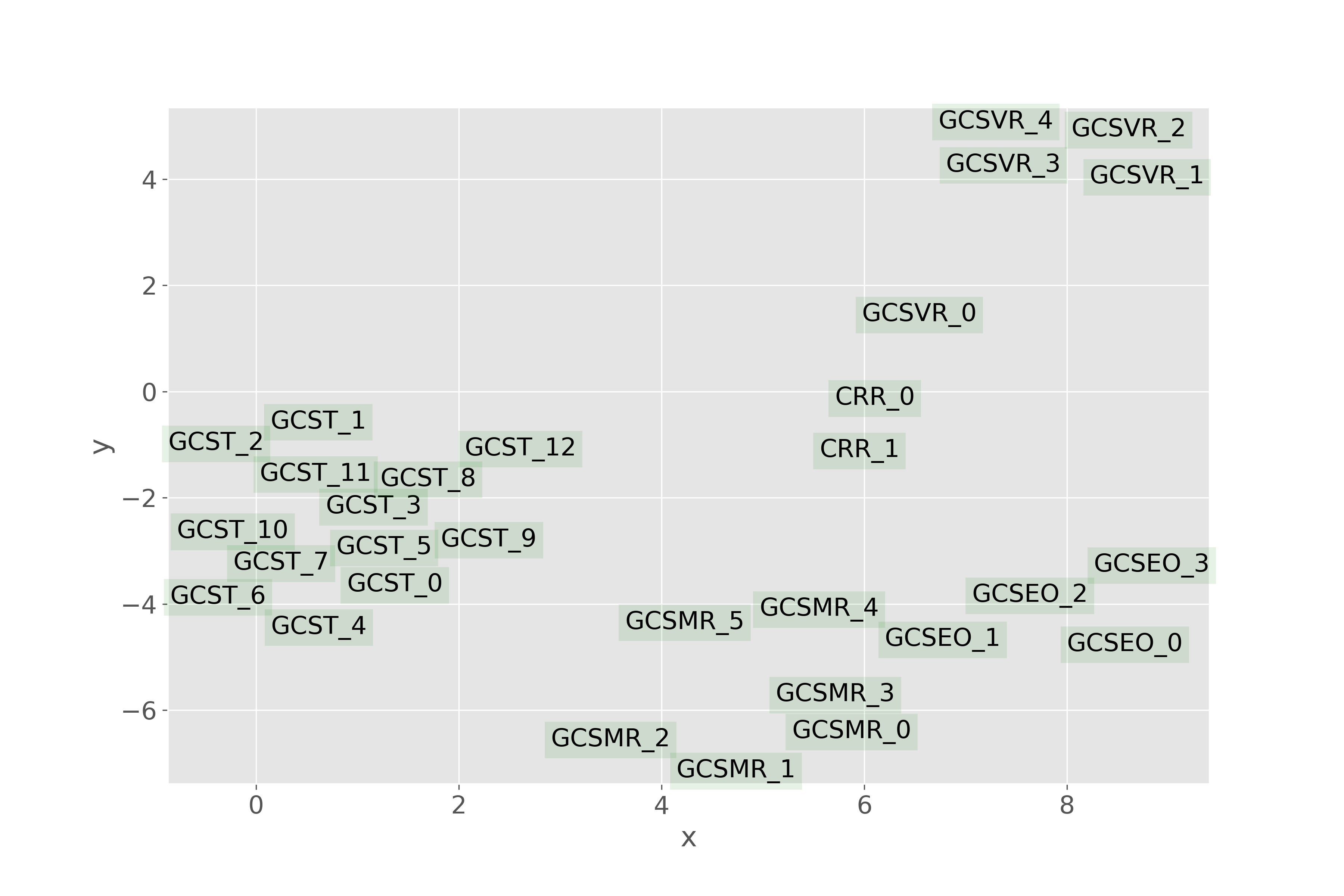}
    \caption{\textbf{FTT}}
    \label{fig:tsne_cat_ftt}
    \end{subfigure}
    \begin{subfigure}[b]{0.45\linewidth}
    \includegraphics[width=\linewidth]      {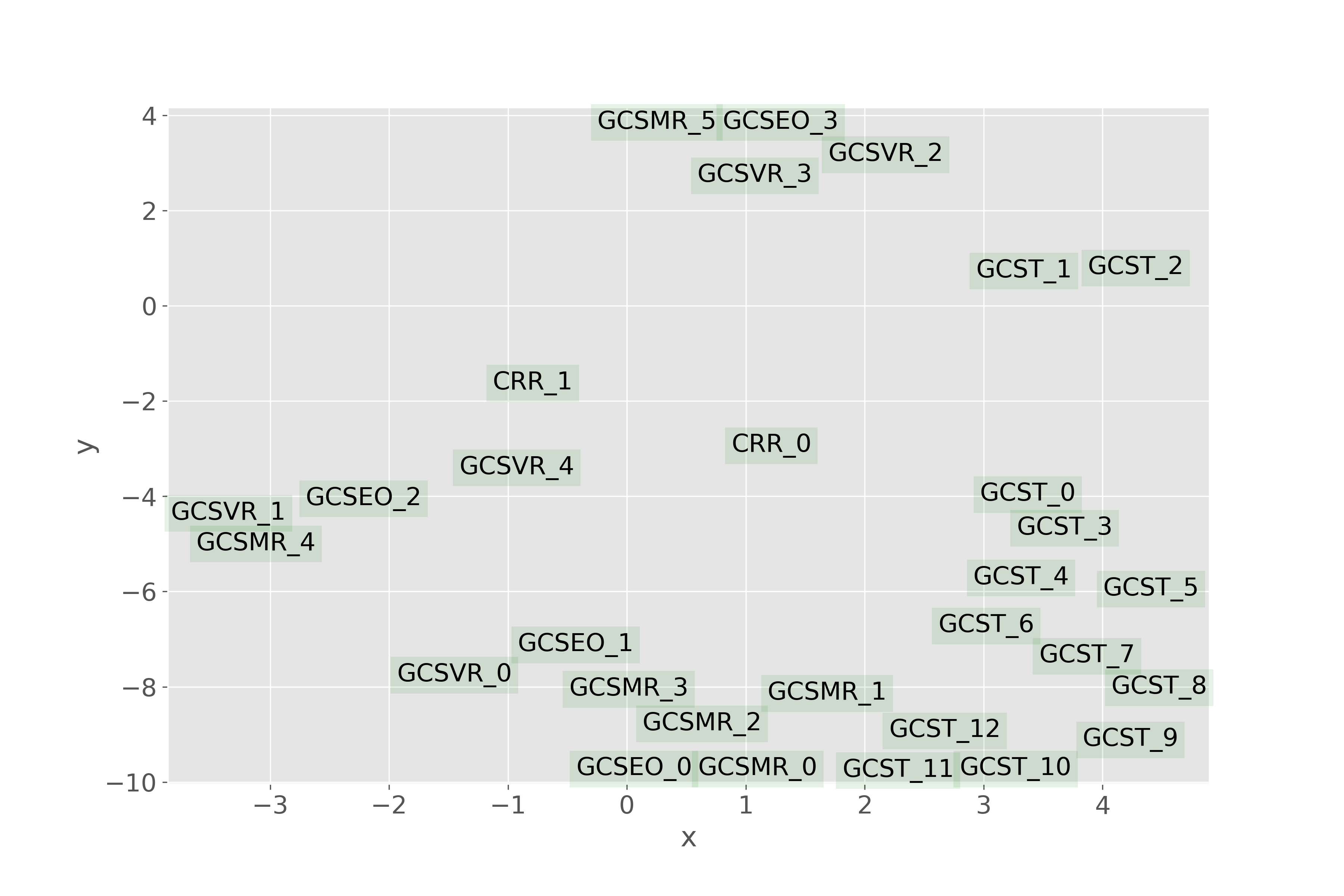}
    \caption{\textbf{CBOW}}
    \label{fig:tsne_cat_cbow}
    \end{subfigure}
    \begin{subfigure}[b]{0.45\linewidth}
    \includegraphics[width=\linewidth]      {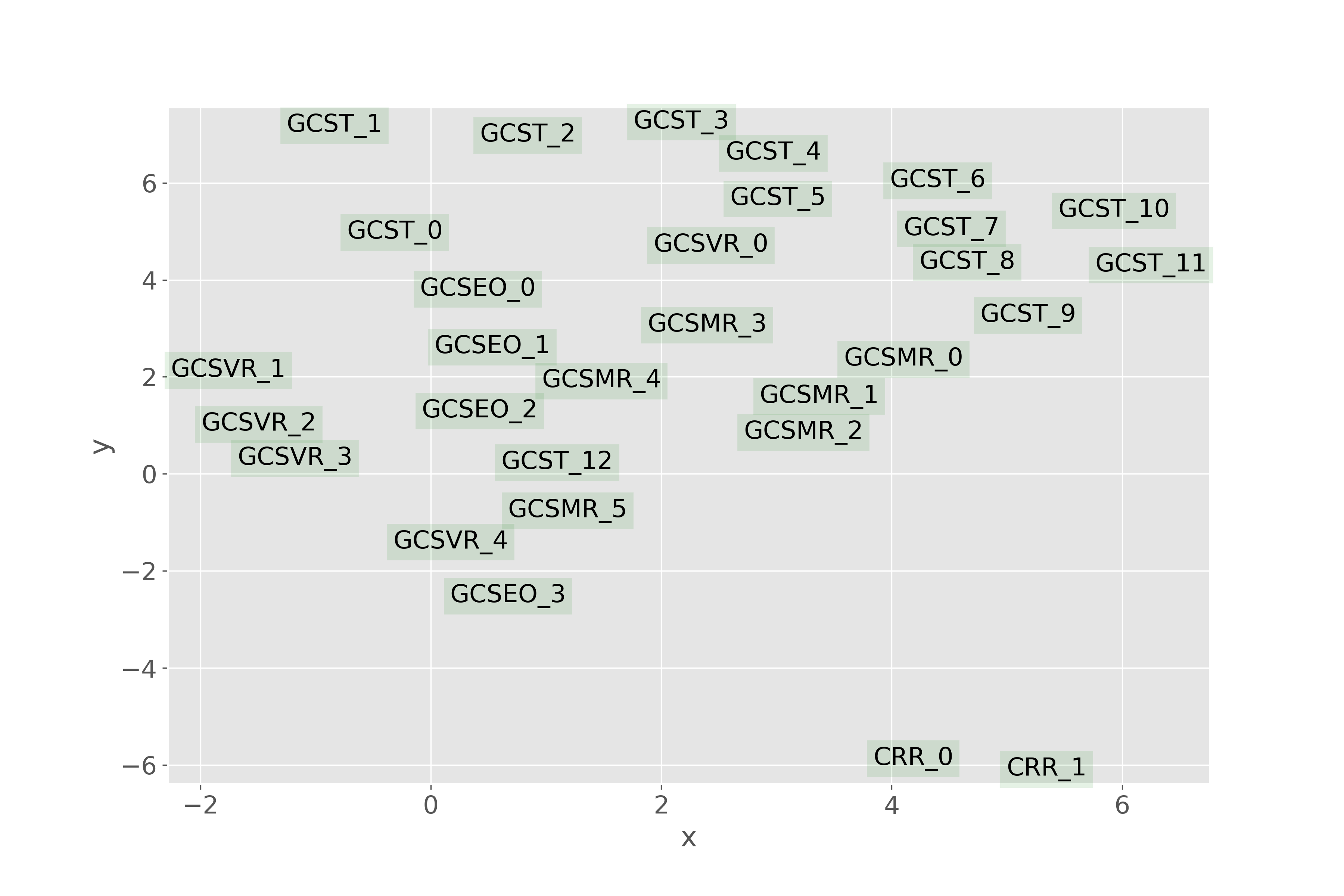}
    \caption{\textbf{MLM}}
    \label{fig:tsne_cat_mlm}
    \end{subfigure}
    \caption{T-SNE visualization, with the perplexity value set to $15$,  of categorical feature embeddings from FTT, CBOW and MLM.}
    \label{fig:tsne_cat}
\end{figure}

\paragraph{Limited Labeled Data Analysis}
To further explore the effect of pre-trained feature embeddings on the decompensation prediction task, we gradually reduce the number of labeled data from $100\%$ to $1\%$ during fine-tuning. The results are shown in Table \ref{tab:limited_label}.
\begin{table}[htbp]
   \centering
  \begin{tabular}{cccc}
    \toprule[2pt]
    Labels & Models & AUPRC & AUROC \\ 
    \hline
    \multirow{4}{*}{$100\%$} & Transformer & $34.4\pm 0.4$& $91.2\pm 0.1$ \\
    & FTT & $36.4\pm 0.2$& $91.6\pm 0.1$\\
    & CBOW & $36.3\pm 0.4$& $91.4\pm 0.1$\\
    & MLM & $36.2\pm 0.1$& $91.6\pm 0.1$\\
    \hline
    \multirow{4}{*}{$50\%$} & Transformer & $33.1\pm 0.1$& $90.8\pm 0.1$ \\
    & FTT & $35.8\pm 0.3$& $91.2\pm 0.1$\\
    & CBOW & $34.8\pm 0.2$& $91.1\pm 0.1$\\
    & MLM & $34.0\pm 0.5$& $91.1\pm 0.1$\\
    \hline
    \multirow{4}{*}{$10\%$} & Transformer & $31.2\pm 0.2$& $90.0\pm 0.1$ \\
    & FTT & $31.3\pm 0.3$& $89.8\pm 0.5$\\
    & CBOW & $28.9\pm 0.6$& $88.6\pm 0.2$\\
    & MLM & $31.3\pm 0.3$& $89.4\pm 0.1$\\
    \hline
    \multirow{4}{*}{$1\%$} & Transformer & $22.2\pm 1.0$& $85.5\pm 1.0$ \\
    & FTT & $8.7\pm 7.0$& $63.2\pm 18.5$\\
    & CBOW & $19.1\pm 1.8$& $83.6\pm 0.4$\\
    & MLM & $11.2\pm 1.0$& $78.4\pm 2.5$\\
    \bottomrule[2pt]
    \end{tabular}
    \caption{Performance on the decompensation task for different models with decreasing labeled data. Mean and standard deviation are reported over three runs.}
    \label{tab:limited_label}
\end{table}

\paragraph{Ablation Study on CBOW}
Besides the traditional CBOW model, we also explored adding information from the previous time step to predict current feature values. The results are shown in Table \ref{tab:ablation}. It turns out that the adapted CBOW pre-training does not improve the downstream task performance.

\begin{table}[htbp]
    \centering
    \begin{tabular}{ccc}
    \toprule[2pt]
    Use\_previous & AUPRC & AUROC \\
    \hline
    False & $36.3\pm 0.4$& $91.4\pm 0.1$\\
    True & $35.8\pm 0.1$& $91.4\pm 0.1$\\
    \bottomrule[2pt]
    \end{tabular}
    \caption{Comparing the performance of adapted CBOW objectives on decompensation task.}
    \label{tab:ablation}
\end{table}

\paragraph{Additional Explanations}
Although we observe good alignment between learnt clinical feature embeddings and prior clinical knowledge, the performance on the decompensation and mortality prediction tasks is unexpectedly not improved when leveraging the pre-trained embeddings. We believe it is necessary to conduct a series of experiments on various downstream tasks to see whether our pre-trained embeddings can help. Besides, the pre-trained clinical embeddings could be combined with higher-level embeddings (e.g. time-series level or patient-level embeddings) to further improve performances on downstream tasks.

\end{document}